\documentclass[sigconf]{acmart}
\pdfoutput=1

\AtBeginDocument{%
  \providecommand\BibTeX{{%
    \normalfont B\kern-0.5em{\scshape i\kern-0.25em b}\kern-0.8em\TeX}}}

\setcopyright{acmcopyright}
\copyrightyear{2018}
\acmYear{2018}
\acmDOI{10.1145/1122445.1122456}

\acmConference[Woodstock '18]{Woodstock '18: ACM Symposium on Neural
  Gaze Detection}{June 03--05, 2018}{Woodstock, NY}
\acmBooktitle{Woodstock '18: ACM Symposium on Neural Gaze Detection,
  June 03--05, 2018, Woodstock, NY}
\acmPrice{15.00}
\acmISBN{978-1-4503-XXXX-X/18/06}

\usepackage{amsmath}

\DeclareMathOperator*{\argmax}{arg\,max}
\DeclareMathOperator*{\argmin}{arg\,min}

\usepackage{multirow}
\usepackage{graphicx}
\usepackage{amsmath}

\usepackage{bbm}

\usepackage{commath}

\usepackage{enumitem}
\usepackage{fixltx2e}

\usepackage[ruled,vlined]{algorithm2e}

\usepackage{romannum}

\usepackage{graphicx}

\usepackage{subcaption}
\usepackage{cleveref}

\begin{document}

\title{SPMoE: Generate Multiple Pattern-Aware Outputs \\with Sparse Pattern Mixture of Experts}

\author{Shaobo Cui}
\affiliation{%
  \institution{DAMO Academy, Alibaba Group}
  \streetaddress{}
  \city{}
  \country{}}
\email{yuanchun.csb@alibaba-inc.com}

\author{Xintong Bao}
\affiliation{%
  \institution{DAMO Academy, Alibaba Group}
  \streetaddress{}
  \city{}
  \country{}}
\email{xintong.bxt@alibaba-inc.com}

\author{Xuming Lin}
\affiliation{%
  \institution{DAMO Academy, Alibaba Group}
  \streetaddress{}
  \city{}
  \country{}}
\email{xuming.lxm@alibaba-inc.com}

\author{Zhongzhou Zhao}
\affiliation{%
  \institution{DAMO Academy, Alibaba Group}
  \streetaddress{}
  \city{}
  \country{}}
\email{zhongzhou.zhaozz@alibaba-inc.com}

\author{Ji Zhang}
\affiliation{%
  \institution{DAMO Academy, Alibaba Group}
  \streetaddress{}
  \city{}
  \country{}}
\email{zj122146@alibaba-inc.com}

\author{Wei Zhou}
\affiliation{%
  \institution{DAMO Academy, Alibaba Group}
  \streetaddress{}
  \city{}
  \country{}}
\email{fayi.zw@alibaba-inc.com}

\author{Haiqing Chen}
\affiliation{%
  \institution{DAMO Academy, Alibaba Group}
  \streetaddress{}
  \city{}
  \country{}}
\email{haiqing.chenhq@alibaba-inc.com}


\begin{abstract}
Many generation tasks follow a one-to-many mapping relationship: each input could be associated with multiple outputs. Existing methods like Conditional Variational AutoEncoder~(CVAE) employ a latent variable to model this one-to-many relationship. However, this high-dimensional and dense latent variable lacks explainability and usually leads to poor and uncontrollable generations.
In this paper, we innovatively introduce the linguistic concept of pattern to decompose the one-to-many mapping into multiple one-to-one mappings and further propose a model named Sparse Pattern Mixture of Experts~(SPMoE). 
Each one-to-one mapping is associated with a conditional generation pattern and is modeled with an expert in SPMoE.  
To ensure each language pattern can be exclusively handled with an expert model for better explainability and diversity, a sparse mechanism is employed to coordinate all the expert models in SPMoE. 
We assess the performance of our SPMoE on the paraphrase generation task and the experiment results prove that SPMoE can achieve a good balance in terms of quality, pattern-level diversity, and corpus-level diversity. 
\end{abstract}





\maketitle

\section{Introduction}
Many conditional generation tasks like paraphrase generation~\cite{xu2018d}, machine translation~\cite{he2018sequence} and dialogue generation~\cite{zhao2017learning} follow a one-to-many mapping relationship: given a source sequence, there are multiple possible target sequences. 
For instance, there are a variety of paraphrases for a given sentence. 
Existing approaches for this one-to-many generation task can be divided into two types: (1) variational approach~\cite{sohn2015learning}: encoding the variability of multiple outputs into a latent variable. Each possible valid output is collaboratively determined by the source input and a latent variable. The diversity of outputs is achieved with the variability of this latent variable.  (2) Decoding strategy: adopting beam search or diverse beam search~\cite{vijayakumar2016diverse} to generate diverse outputs in the decoding phase. 
However, these multiple outputs lack the explainability and controllable distinction, i.e., these methods can't make sure that these outputs can distinct from each other and the outputs may lack diversity and are low-quality. 

To better tackle the one-to-many mapping problem, we borrow the concept of \textbf{pattern} from the linguistic community and slightly abuse this concept. 
A one-to-many mapping can be decomposed into multiple one-to-one mappings. Each one-to-one mapping follows a determined pattern. 
To better illustrate the intuition of \textbf{pattern}, we take the task of paraphrase as an instance and list several examples in Table~\ref{tab:illustration_case}. 
Given a source sentence, we could obtain different paraphrase outputs when applied with different paraphrase patterns. 
Each <input, output> pair can be viewed as an instance of a paraphrase pattern or the combination of several patterns. With the introduction of the concept of pattern, the knotty one-to-many mapping of conditional generation task can be converted as multiple pattern mappings, each of which is a determined one-to-one mapping. 
\begin{table}[]
    \centering
    \small 
    \resizebox{0.45\textwidth}{!}{
    \begin{tabular}{p{3.2cm}|p{5cm}}
       \toprule
      \textbf{}   & \textbf{Text} \\ \hline 
        {Source sentence} & John failed in the final Physics exam this semester due to his carelessness.  \\ \hline
        \textit{Synonyms replace pattern} & John failed in the final Physics exam this term due to his carelessness.  \\ 
        \textit{Form change pattern} & John's failure in the final Physics exam this term is due to  his carelessness.  \\ 
        \textit{Grammatical change pattern} & Since he is careless, John failed in the final Physics exam this semester.\\
        \textit{Combined Patter}n & Since he is careless, John failed in the final Physics exam this term. \\ 
    \bottomrule
    \end{tabular}
    }
    \caption{An example case for paraphrase pattern}
    \label{tab:illustration_case}
\end{table}

To capture the aforementioned intuition, we propose to model the one-to-many mapping with a model named Sparse Pattern Mixture of Experts~(SPMoE). 
There are multiple expert models in SPMoE. Each expert is expected to \textbf{exclusively} models a unique one-to-one mapping pattern, i.e., each expert model should distinguish itself from other expert models. 
To push each expert to be as distinguishable as possible, each expert model should be provided data samples with different distributions. Namely, each data sample should be assigned to an expert model~(we can relax this only one expert to a limited number of expert models). In other words, the probability of each sample being classified to the mixture of experts is expected to be a sparse probability simplex.
For this end, we introduce a sparse transformation mechanism~\cite{martins2016softmax} to enforce the distribution to be sparse. In other words, the sparse mechanism encourages each sample can be \textbf{exclusively} owned by a certain expert model.
In addition, to avoid the \textit{the rich become richer} phenomenon~\footnote{Only one expert get trained if it is slightly better than others while other expert models are ignored~\cite{shen2019mixture}}, we incorporate a batch-level load balance strategy in the sparse mechanism to encourage all the experts to be well-involved in the pattern training process. 
The batch-level load balance collaborates with the instance-level sparse transformation to achieve two desirable properties: 
(1) Each expert model in SPMoE can learn the inexplicit pattern in an unsupervised fashion and each expert is assured to be well-involved with the batch-level load balance strategy. 
(2) Each expert model can well distinguish from others with the enforcement of sparse transformation.

Furthermore, we propose a metric dubbed Pattern Diversity~(PD) to evaluate the diversity among the multiple outputs for a given input. Compared with existing metrics like Pairwise-BLEU~\cite{shen2019mixture} or Self-BLEU~\cite{yu2017seqgan}, PD enforces a sentence brevity penalty term and enjoy better explainability for n-gram diversity. 
Our approach is general and can easily be applied to existing conditional generation tasks. We conduct our experiment on the classical paraphrase task and the experimental results prove that the SPMoE model can achieve a satisfying trade-off between quality and diversity. 
To sum up, our contributions are as follows:
\begin{itemize}[leftmargin=*]
    \item We model the one-to-many generation task from the perspective of linguistic pattern and further propose a novel framework named SPMoE to model this one-to-many mapping.  
    \item We incorporate the sparse transformation and batch-level load balance mechanism to encourage multiple expert models in SPMoE can distinguish each other while still being well-involved. 
    \item We propose a novel metric: Pattern Diversity~(PD) to evaluate the diversity among these multiple outputs for a given input, which enjoys a better explainability and enforces a sentence brevity penalty term. 
\end{itemize}

\begin{figure}
    \centering
    \includegraphics[width=0.95\columnwidth]{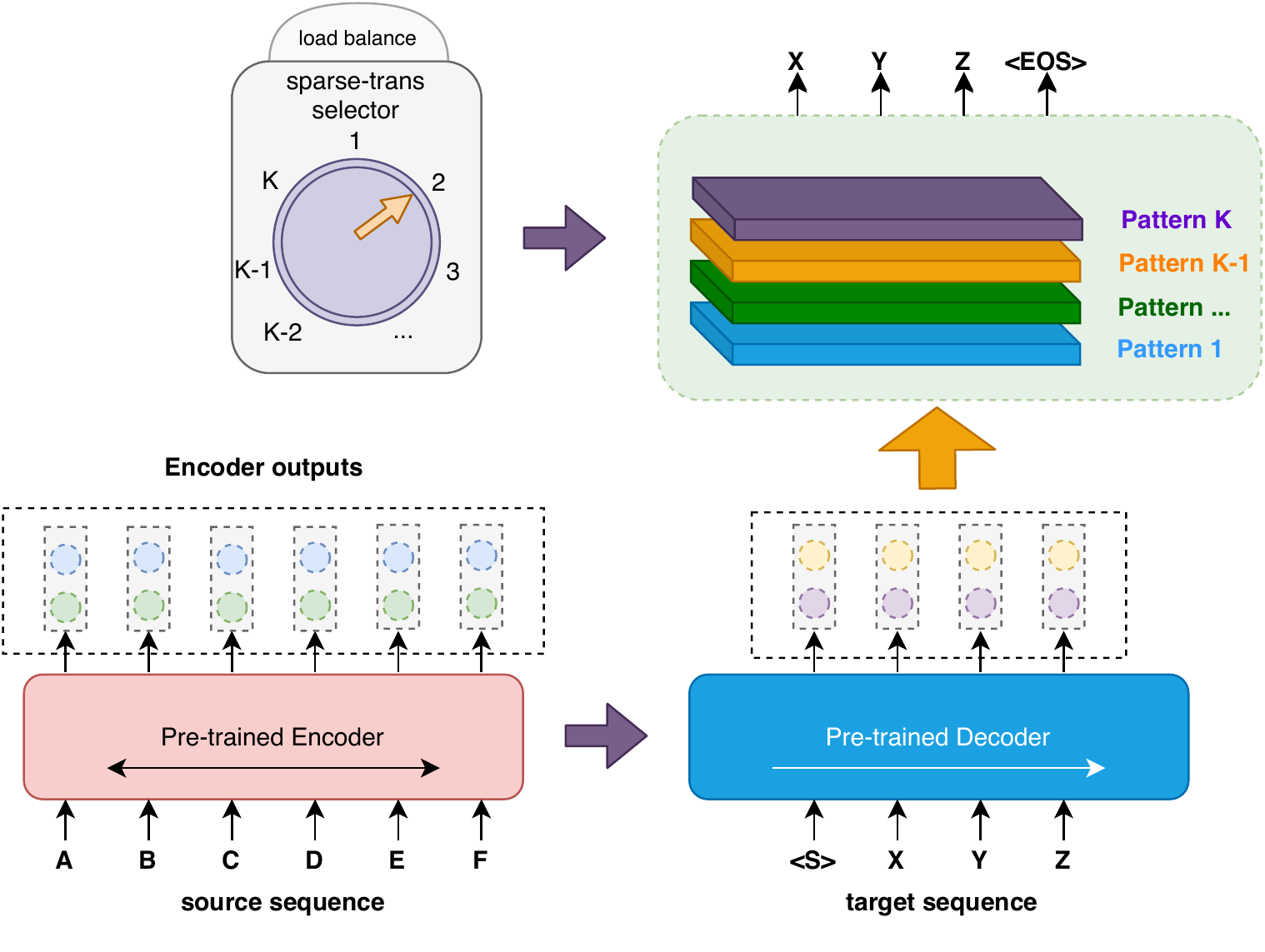}
    \caption{Framework of SPMoE}
    \label{fig:model}
\end{figure}

\section{Model}
\subsection{Overview}
For a given corpus: $C = \{(x_1, y_1), (x_2, y_2), \cdots, (x_n, y_n) \}$, $x_i$ is the source sentence while $y_i$ is the output sentence. 
Suppose that there are $K$ generation patterns: $F = \{F_1, F_2, \cdots, F_K\}$ in $C$. Each generation pattern can map the source sequence into corresponding pattern-specific target sequence. 
Our goal is to automatically extract the patterns from the one-to-one mapping corpus $C$. In the generation process, we can iterate all the patterns to generate pattern-aware outputs. 
Formally,     
\begin{equation}
        y_k = \argmax_{y} p(y \vert x, F_k).
\end{equation}

We present the model overview of SPMoE in Figure~\ref{fig:model}. 
The backbone of SPMoE is a transformer-based~\cite{vaswani2017attention} model which can be easily adapted from pre-trained language models like BART~\cite{lewis2020bart}. All expert models in SPMoE share the same backbone structure while their pattern heads are separate and are implemented with a linear layer. 
All the pattern heads are coordinated with a sparse mechanism and a load balance strategy to enforce each expert model to be as distinct as possible. 
Specifically, the backbone is used for a better representation of the source-target pairs while the sparse mechanism is used to allocate these pairs into different patterns. 

\subsection{Pattern Clustering with Sparse Mechanism} \label{sec:model:sparse}
\noindent\textbf{Representation of Source-Target Pairs} \quad
With the encoder and decoder structure of pre-trained mode, we use the encoder outputs $\mathbf{E}_{\text{enc}}$ and decoder outputs $\mathbf{E}_{\text{dec}}$ as the representation of source sentence and target sentence. The overall representation is defined as: 
\begin{align}
    z^{\text{pattern}} =  [\mathbf{E}_{\text{enc}} \cdot \mathbf{w}_{\text{enc}}; \mathbf{E}_{\text{dec}} \cdot \mathbf{w}_{\text{dec}}] \cdot \mathbf{w} + \mathbf{b},
\end{align}
where $\mathbf{w}_{\text{enc}} \in \mathbb{R}^{L_{\text{input}}}$, $\mathbf{w}_{\text{dec}} \in \mathbb{R}^{L_{\text{output}}}$, $L_{\text{input}}$ and $L_{\text{input}}$ are the length of input utterance and output utterance. $\mathbf{w} \in \mathbb{R}^{2d_{\text{model}} \times K}$, $\mathbf{z}^{\text{pattern}} \in \mathbb{R}^{K}$. 

\noindent\textbf{Sparse Transformation: Sparsegen-lin} \quad 
Then a sparse transformation is adopted to enforce sparse pattern selection. This sparse transformation is abstracted as:
\begin{equation}
    \mathbf{p}^{\text{pattern}} = \text{sparse-trans}(\mathbf{z}^{\text{pattern}}),
\end{equation}
where $\mathbf{p}^{\text{pattern}}$ is a point in $(K-1)$-dimensional probability simplex: $\Delta^{K-1}$~\footnote{$\Delta^{K-1} \coloneqq \{\mathbf{p} \in \mathbb{R}^{K} \vert \mathbf{1}^\top \mathbf{p} = 1, \mathbf{p} > 0 \}$}. 
We are interested in maps from $\mathbb{R}^K$ to $\Delta^{K-1}$. 
We want the map function to have the following properties: 
(1) This map function should be differential. 
(2) This map function should lead to a sparse solution. In this case, each expert model can exclusively own the samples. This makes every expert model can cluster the samples following different patterns, which will in return make each expert model can distinguish from each other. 
With the aforementioned requirements, we adopt sparsegen-lin~\cite{laha2018controllable} as the sparse transformation function: 
\begin{equation}
\begin{split}
    \text{sparsegen-lin}(\mathbf{z}; \lambda) &\coloneqq \argmin_{\mathbf{p} \in \Delta^{K-1}} {\Vert \mathbf{p} - \mathbf{z} \Vert}_2^2 - \lambda {\Vert\mathbf{p}\Vert}^2_2  \\
    &= \argmin_{\mathbf{p} \in \Delta^{K-1}} \sum^{K}_{i}(1 - \lambda) p^2_i - 2 p_i z_i + z^2_i \\
    &= \argmin_{\mathbf{p} \in \Delta^{K-1}} \sum^{K}_{i} p^2_i - \frac{2}{1 - \lambda} p_i z_i + \frac{z^2_i}{1 - \lambda} \\
    &= \argmin_{\mathbf{p} \in \Delta^{K-1}} \sum^{K}_{i} p^2_i - \frac{2}{1 - \lambda} p_i z_i + {(\frac{z_i}{1 - \lambda})}^2 +  \frac{z^2_i}{1 - \lambda} - {(\frac{z_i}{1 - \lambda})}^2 \\
    &= \argmin_{\mathbf{p} \in \Delta^{K-1}} {\Vert \mathbf{p} - \frac{\mathbf{z}}{1-\lambda} \Vert}_2^2
\end{split}
\label{eq:sparselin}
\end{equation}
where $\lambda \in (-\infty, 1)$ is the hyper-parameter to control the extent of sparsity. The closer $\lambda$ is to 1, the sparser the output distribution will be. Namely, sparsegen-lin returns the Euclidean projection of the input vector $\mathbf{p}$ onto the probability simplex with a regularization term to further encourage sparsity.

\noindent \textbf{Closed-Form Solution of Sparsegen-lin\quad}
Sparsegen-lin has a closed-form solution. The Lagrangian of the optimization problem in Equation~(\ref{eq:sparselin}) is: 
\begin{equation}
    \mathcal{L}(\mathbf{p}, \mathbf{u}, \tau) = \frac{1}{2} {\Vert \mathbf{p} - \frac{\mathbf{z}}{1-\lambda} \Vert}_2^2 - \mathbf{u}^{\top} \mathbf{p} + \tau(\mathbf{1}^{\top} \mathbf{p} - 1) 
\end{equation}
The optimal ($\mathbf{p}^{*}$, $\mathbf{u}^{*}$, $\tau$) should follow the following Karush-Kuhn-Tucker conditions: 
\begin{align}
    \nabla_{\mathbf{p}} \mathcal{L} = \mathbf{p}^{*} - \frac{\mathbf{z}}{1 - \lambda} - \mathbf{u}^{*} + \tau^{*} = \mathbf{0} & \label{eq:stationary}\\
    \mathbf{1}^{\top} \mathbf{p}^{*} - 1 = 0 \label{eq:constraint1}&\\
    \mathbf{p}^{*} \geq \mathbf{0} &\\
    \mathbf{u}^{*} \geq \mathbf{0} & \\
    {u}^{*}_i p^{*}_i = 0, \forall i \in [K] \label{eq:slackness}
\end{align}
We can see that for $j \in [K]$ and ${p}^{*}_{j} > 0$, with the constraint of Equation~(\ref{eq:slackness}), $u^{*}_{j} = 0$. In this setting, with the constraint of Equation~(\ref{eq:stationary}), we have: 
\begin{equation}
    p^{*}_{j} = \frac{z_j}{1 - \lambda} - \tau^{*}.
\end{equation}
Let $S(z) = \{j \in [K] \vert p^{*}_{j} > 0\}$. From Equation~(\ref{eq:constraint1}), we have: 
\begin{equation}
    \sum_{j \in S(z)} (\frac{z_j}{1 - \lambda} - \tau^{*}) = 1
\end{equation}
In this setting, we have: 
\begin{equation}
    \tau^{*} = \frac{\sum_{j \in S(z)} (\frac{z_j}{1 - \lambda}) - 1}{\vert S(\mathbf{z})\vert}
\end{equation}
From Equation~(\ref{eq:slackness}), it is obvious that for $u^{*}_{i} > 0$ implies $p^{*}_{i} = 0$, i.e., $i \notin S(\mathbb{z})$,
which from Equation~\ref{eq:stationary} implies that 
\begin{equation}
\begin{split}
        u^{*}_i &= p_i - \frac{z_i}{1 - \lambda} + \tau^{*} \\
                &=  \tau^{*}  - \frac{z_i}{1 - \lambda} > 0
\end{split}
\label{eq:constaint:feasible}
\end{equation}
Let $z_{(1)} \geq z_{(2)} \geq z_{(3)} \geq \cdots \geq z_{(K)}$. Suppose $k(z) \in S(\mathbf{z})$ while $k(z) + 1 \notin S(\mathbf{z})$.  We can see that $k(z) = \vert S(z) \vert$
\begin{equation}
\begin{split}
    k(z) &\coloneqq  \max \{ k \in [K] \vert \tau^{*} < \frac{z_{k}}{1 - \lambda} \} \\
    &\coloneqq  \max \{ k \in [K] \vert 
    \frac{\sum_{j \leq k} (\frac{z_{(j)}}{1 - \lambda}) - 1}{k}
    < 
    \frac{z_{(k)}}{1 - \lambda} \} \\
    &\coloneqq \max \{ k \in [K] \vert 
    {\sum_{j \leq k} (\frac{z_{(j)}}{1 - \lambda}) - 1}
    < 
    k \cdot \frac{z_{(k)}}{1 - \lambda} \}
\label{eq:kz:requirement}
\end{split}
\end{equation}
See more about sparsemax and sparsegen-lin in \cite{peters2019sparse,martins2016softmax}. 

\subsection{Reconstruction Loss}
Given a source-target pair $(x,y)$, the cross entropy loss associated with these K patterns can be denoted as $\mathbf{CE} \in \mathbb{R}^{K}$. The $k_{th}$ element in $\mathbf{CE}$ is calculated as: 
\begin{equation}
    CE_k(x, y) = -\sum^{L}_{t=1} \log prob(y_t \vert y_{<t}, x; \theta, h_k),
\end{equation}
where $h_k$ is the pattern head associated with pattern $k$ and $\theta$ represents the backbone model shared by all the patterns. With the pattern distribution $\mathbf{p}$ and $K$ generation patterns: $F = \{F_1, F_2, \cdots, F_K\}$, we have: 
\begin{equation}
\begin{split}
    prob(y \vert x, F, \mathbf{p}) &= \sum_{j \in [\vert F \vert]} \mathbf{p}_j \cdot prob(y \vert x, F_j) \\
    &= \sum_{j \in [\vert F \vert]} \mathbf{p}_j \cdot \prod^{L}_{t=1} prob(y_t \vert y_{<t}, x, h_j; \theta) \\
    &= \sum_{j \in [\vert F \vert]} \mathbf{p}_j \cdot \frac{1}{\exp{(CE_j(x, y))}}
\end{split}
\end{equation}

The reconstrunction loss can be expressed as: 
\begin{equation}
\begin{split}
    L_{\text{rec}} &= - \log prob(y \vert x, F, \mathbf{p}) \\
    &= - \log \sum_{j \in [\vert F \vert]} \mathbf{p}_j \cdot \frac{1}{\exp{(CE_j(x, y))}} \\
\end{split}
\end{equation}

\subsection{To Prevent the Rich From Becoming Richer: Load Balance Loss}
From our experiment, we observe that only invariable few pattern heads are selected and trained while others are seldom optimized. This will lead to the \textit{rich become richer} phenomenon~\cite{shen2019mixture}. 
Once a pattern head is slightly better than others, it is more likely to be picked and optimized. 
To implicitly enforce the load balance of pattern heads, i.e., each pattern head can be picked almost the same number of times, inspired by \cite{shazeer2017outrageously}, we define an addition loss $L_{\text{balance}}$: 
\begin{equation}
    L_{\text{balance}} = KL(\frac{\sum_{i \in B} \mathbf{p}^{i}}{\vert B \vert} \Vert U),
\end{equation}
where $KL(P \Vert Q) = \sum_{i} P(i) \ln \frac{P(i)}{Q(i)}$ is the Kullback-Leibler divergence and $U$ is the uniform distribution. 
The final loss is the combination of reconstruction loss and load balance loss: 
\begin{equation}
    L_{\text{final}} = L_{\text{rec}} + \gamma L_{\text{balance}}
\end{equation}
In this way, SPMoE has two desired properties: 
\begin{itemize}
    \item Each expert model is pushed to be as distinct as possible with the sparse mechanism for pattern selection, which in return ensures the output utterances generated by each expert model to be pattern-specific. 
    \item With the load balance loss, each pattern head can be picked almost the same number of times. This makes sure that each expert is involved and well-trained, which makes the output utterance generated by each expert model is high-quality.
\end{itemize}

\section{PD: a Method for Automatic Evaluation of Pattern-level Diversity}
We expect to measure the difference and diversity among multiple outputs for a given input. There are several existing metrics for pattern-level diversity such as Pairwise-BLEU and Self-BLEU. These metrics are motivated by the fact that a higher BLEU means a higher resemblance between two sentences, and thus a lower diversity. However, we find the fact that under this type of metrics, one long sentence and another sentence contain a single word from the first sentence could achieve a satisfying pattern diversity score since the BLEU score between them is quite low, which counters our intuition. Motivated by the aforementioned shortcomings, we propose a metric named \textit{Pattern Diversity}~(PD). 
PD is based on n-gram diversity and applied with sentence brevity penalty. 
\subsection{$n$-gram Pattern Diversity}
The n-gram pattern diversity is defined as: 
\begin{equation}
    diver_{n} = \frac
{\sum\limits_{\textit{n-gram} \in \mathcal{C}} Count(\textit{n-gram}) - \sum\limits_{\textit{n-gram} \in \mathcal{R}} Count_{clip}(\textit{n-gram})}
{\sum\limits_{\textit{n-gram} \in \mathcal{C}} Count(\textit{n-gram})},
\end{equation}
where $Count(n-gram)$ is the number of this $n$-gram in the candidate sentence. $\mathcal{C}$ is the candidate sentence and $\mathcal{R}$ is the reference sentence. And similar as the famous BLEU~\cite{papineni2002bleu}, we adopt the clip strategy: 
\begin{equation}
    Count_{clip}(n-gram) = \min (Count(n-gram),\text{Ref\_Count(n-gram)}),
\end{equation}
where Ref\_Count(n-gram) is the count of n-gram observed in the reference sentence. 

In this way, we truncate each word's count. We give an example in Example 1. 

\vspace{0.1cm}
\noindent\fbox{%
    \parbox{0.3\textwidth}{%
\noindent\textbf{Example 1} \\
\noindent Candidate: {\textsf{The the the the the the the.}} \\
\noindent Reference: {\textsf{\underline{The} cat is on \underline{the} mat.}}
}
}
\vspace{0.1cm}

With the clip operation, the modified $1$-gram diversity is:
\begin{equation}
    diver_1 = \frac{7 - 2(the) - 0(cat) - 0(is) - 0(on) - 0(mat)}{7} = \frac{5}{7}. 
\end{equation}
Without the clip operation, the $1$-gram diversity is computed as: 
\begin{equation}
    diver_1 = \frac{7 - 7(the)- 0(cat) - 0(is) - 0(on) - 0(mat)}{7} = 0
\end{equation}

\subsection{Sentence Length Mismatch Penalty} \quad In our setting, we expect the candidate to have as many novel tokens as possible. As we can see, the candidate in Example 1 has modified unigram diversity $diver_{1} = \frac{1}{1} = 1$, which contradicts to our common sense. To penalize this type of cases, we need sentence brevity. 

\vspace{0.1cm}
\noindent\fbox{%
    \parbox{0.3\textwidth}{%
\noindent\textbf{Example 2} \\
\noindent Candidate(\textit{Pattern A}): {\textsf{Dog. }} \\
\noindent Reference(\textit{Pattern B}): {\textsf{The cat is on the mat.}}
    }
}
\vspace{0.1cm}

Following \cite{papineni2002bleu}, we adopt a sentence brevity penalty term BP: 
\begin{equation}
\text{BP} = 
\begin{cases}
1 \qquad & \text{if} \quad c > r \\
e^{1 - \frac{r}{c}} & \text{if} \quad c \leq r
\end{cases},
\end{equation}
where $c$ and $r$ represents the sentence length of the candidate sentence and reference sentence. 

\subsection{Pattern Diversity} 
We first compute the geometric average of the modified $n$-gram diversity, $diver_n$, using $n$-grams up to length $N$ and positive weights $w_n$ summing to one. 
Then this geometric average is multiplied by an exponential brevity penalty factor: 
\begin{equation}
    \text{PD-N} = \text{BP} \cdot \exp{\Large(\sum\limits^{N}_{n=1} w_n \log(diver_n)\Large)} 
\end{equation}

\section{Experiments}

\begin{table*}[!tp]
  \centering
  \caption{The comparison of baselines on paraphrase generation.}
    \resizebox{0.95\textwidth}{!}{
    \begin{tabular}{lrrrrrrrrrrrrrrrr}
    \toprule
    \multirow{2}{*}{Models} & \multicolumn{7}{c}{Similarity with ground-truth} & \multicolumn{8}{c}{Pattern-level diversity} & \multicolumn{1}{c}{Corpus-level diversity}\\
    \cmidrule(lr){2-8} \cmidrule(lr){9-16} \cmidrule(lr){17-17}
     & BLEU-1 & BLEU-2 & BLEU-3 & BLEU-4 & Rouge-1 & Rouge-2 & Rouge-L & P-BELU-1~$\downarrow$ & P-BLEU-2~$\downarrow$ & P-BLEU-3~$\downarrow$ & P-BLEU-4~$\downarrow$ & PD-1 & PD-2 & PD-3 & PD-4 & DISTINCT-1\\ 
    \midrule
    {VAE}   & 0.536 & {0.443} & {0.378} & {0.328} & {0.613} & {0.431} & {0.614} & 0.677 & 0.572 & 0.484 & 0.412 & 0.296 & 0.369 & 0.423 & 0.466 & 0.013\\ 
    {CVAE}   & 0.452 & 0.300 & 0.216 & 0.162 & 0.481 & 0.210 & 0.475 & 0.478 & 0.319 & 0.218 & 0.150 & 0.413 & 0.508 & {0.569} & {0.612} & 0.013\\
    {T-CVAE}   & 0.307 & 0.220 & 0.165 & 0.128 & 0.342 & 0.166 & 0.332 & {0.382} & {0.247} & {0.175} & {0.126} & {0.450} & {0.517} & {0.564} & 0.598 & {0.036} \\
    DBS & 0.528 & 0.409 & 0.325 & 0.263 & 0.554 & 0.328 & 0.543 & 0.789 & 0.736 & 0.687 & 0.640 & 0.169 & 0.205 & 0.234 & 0.259 & 0.025 \\
    Sampling   & {0.551} & 0.429 & 0.343 & 0.280  & 0.564 & 0.336 & 0.552 & 0.976 & 0.971 & 0.967 & 0.960 & 0.029 & 0.033 & 0.037 & 0.040 & 0.021 \\
    \midrule 
    SPMoE  & 0.517 & 0.395 & 0.312 & 0.252 & 0.519 & 0.300 & 0.508 & 0.692 & 0.643 & 0.594 & 0.548 & 0.216 & 0.250 & 0.279 & 0.302 & 0.024\\
    \bottomrule
    \end{tabular}
  }
  \label{tab:paraphrase}
\end{table*}

\subsection{Experimental Setup}
We evaluate the involved baselines on Quora question paraphrase dataset~\footnote{ https://www.kaggle.com/c/quora-question-pairs}. However, our method is also suitable for other conditional generation tasks. 
We evaluate the generation outputs in terms of quality and diversity. The quality is evaluated by BLEU~\cite{papineni2002bleu} and Rouge~\cite{lin2004rouge}. 
We measure the diversity from two aspects: 
(1) Corpus-level diversity: measuring the overall diversity for generations of all the input sentences. We use DISTINCT~\cite{li2016diversity} as the metric to reflect the diversity of generation at the corpus level. 
(2) Pattern-level diversity: we measure the difference and diversity among multiple outputs for a given input with Pairwise-BLEU and our proposed PD.

\subsection{Baselines}
\begin{itemize}[leftmargin=*]
    \item VAE: a variational autoencoder~(VAE) model~\cite{gupta2018deep} that conditions the encoder and decoder module of VAE on the source sentence. 
    \item CVAE: a conditional variational autoencoder~(CVAE) model~\cite{zhao2017learning} that encodes the input sentence into a latent variable while the decoder generates the output with the latent variable and the source sentence. 
    \item T-CVAE: a conditional variational autoencoder~(CVAE) model that is implemented based on transformers.  
    \item Sampling: generate outputs with the sampling decoding strategy by a finetuned BART model~\cite{lewis2020bart}. 
    \item Diverse Beam Search~(DBS): generate outputs with diverse beam search decoding strategy~\cite{vijayakumar2016diverse} by a finetuned BART model~\cite{lewis2020bart}, which decodes diverse lists via dividing the beam budget into several groups and encourages diversity between groups with a diversity penalty term. 
\end{itemize}
\subsection{Experimental Results}
We present the results of paraphrase generation in Table~\ref{tab:paraphrase}. We have the following observations: (1) VAE model~\cite{gupta2018deep} achieve the best performance on the similarity to the ground truth, which even surpasses the performance of BART-based methods~(Sampling and DBS). However, its corpus-level diversity is the poorest. 
(2) T-CVAE and CVAE achieve the best performance in terms of pattern-level diversity. This means that latent variable is useful for diverse content generation. However, CVAE and T-CVAE are quite poor in terms of quality, i.e., the generated output has a low BLEU score.
(3) T-CVAE is the best model on corpus-level diversity. 
(4) Our SPMoE model achieves a good balance among the quality, pattern-level diversity and corpus-level diversity. Additionally, same as Sampling and DBS, SPMoE can be easily modified from existing pre-trained language models. 

\section{Conclusions}
In this paper, we introduce the concept of pattern to decompose the one-to-many generation problem into multiple one-to-one mappings and further propose a sparse pattern mixture of experts model to produce diverse and high-quality generation. 
In the future, we plan to extend SPMoE to more generation tasks and utilize the image generation task to verify the effectiveness of SPMoE.

\bibliographystyle{ACM-Reference-Format}
\bibliography{sample-base}

\end{document}